\documentclass[twocolumn]{svjour3}          
\smartqed  
\usepackage{graphicx}
\usepackage{amsmath}
\usepackage{booktabs}
\usepackage{url}
\usepackage{longtable}
\usepackage{multirow,colortbl}
\usepackage{titlesec}
\usepackage{adjustbox}
\usepackage[table,xcdraw]{xcolor}
\begin{document}

\title{Content-based jewellery item retrieval using the local region-based histograms}

\author{Amin Muhammad Shoib$^{1\textbf{*}}$ \and Jabeen Summaira$^{2}$ \and Changbo Wang$^{3}$  \and Ali Tassawar$^{4}$
}


\institute{  	$^{1}$\textbf{*} Muhammad shoib Amin\\
				\email{52184501030@stu.ecnu.edu.cn}\\ 
				Summaira Jabeen\\
				\email{11821129@mail.zju.edu.cn}\\     
				Changbo Wang\\
				\email{cbwang@sei.ecnu.edu.cn}\\     
				Tassawar Ali\\
				\email{tassawar.ali@gmail.com}\\       \\
				$^{1}$ \at School of Software Engineering,
              	East China Normal University,
              	Shanghai, China \\
                \and
                $^{2}$
          		\at College of Computer Science and Technology, Zhejiang University, Hangzhou, China\\
               	\and
		        $^{3}$
         		\at School of Computer Science and Technology, East China Normal University, Shanghai, China\\
         		\and
         		$^{4}$\at Department of Computer Science, COMSATS University Islamabad, Wah Campus, Pakistan \\          
}

\date{Received: date / Accepted: date}

\maketitle

\begin{abstract}
Jewellery item retrieval is regularly used to find what people want on online marketplaces using a sample query reference image. Considering recent developments, due to the simultaneous nature of various jewelry items, various jewelry goods' occlusion in images or visual streams, as well as shape deformation, content-based jewellery item retrieval (CBJIR) still has limitations whenever it pertains to visual searching in the actual world. 
This article proposed a content-based jewellery item retrieval method using the local region-based histograms in HSV color space. Using five local regions, our novel jewellery classification module extracts the specific feature vectors from the query image. The jewellery classification module is also applied to the jewellery database to extract feature vectors. Finally, the similarity score is matched between the database and query features vectors to retrieve the jewellery items from the database. 
The proposed method performance is tested on publicly available jewellery item retrieval datasets, i.e. ringFIR and Fashion Product Images dataset. 
The experimental results demonstrate the dominance of the proposed method over the baseline methods for retrieving desired jewellery products.

\end{abstract}


\section{Introduction}
\label{intro}
Content-based fashion item retrieval (CBFIR) methods retrieve similar products from a huge cluster of fashion item images. This lessens the dependency of the CBFIR system on text, enabling a better precise and immediate search for the wanted fashion item \cite{shoib2023methods,zhou2019fashion}. CBFIR has many applications in the fashion industry, like fashion item retrievals \cite{alirezazadeh2022deep}, fashion product parsing \cite{jia2023coloration}, fashion outfit recommendations \cite{sarkar2023outfittransformer}, fashion analysis \cite{chen2023survey}, and many more. Many unimodal and multimodal deep learning techniques are used recently for the retrieval of desired fashion products from the huge collection of data streams \cite{jabeen2023review,liao2018interpretable,tautkute2019deepstyle,jo2020flexible,rubio2017multi}. Whenever a customer submits a keyword, source image, or image with some additional text, drawing, or visual stream from their everyday events, the CBFIR may locate fashion items or goods using identical or comparable attributes for e-commerce purchases \cite{amin2022fashion}. Many e-commerce websites provide clothing retrieval services for users by keyword query \cite{cheng2021fashion,zhang2020clothingout}. But keyword-based retrieval can hardly meet the requirement of exact recovery of a particular clothing style. Content-based fashion product retrieval can fill the gap to a certain extent, and some existing websites and applications support content-based fashion item retrieval. But the results of low-level feature-based image retrieval still have semantic gaps with user requirements, and still, there is ample space for improvement in this area. 

In fashion item retrieval systems, desired fashion products from large databases are retrieved using text or reference images. Traditional text-based retrieval systems employ the real text on website information, which simplifies indexing and keyword extraction. Keyword-based or text-based fashion item retrieval techniques have a simpler architecture design when compared to the CBFIR systems \cite{wu2021fashion,chen2020learning}. Prior text-based retrieval systems relied heavily on manual labeling of images, where the experts annotate the portions of images using the specified keywords. Such manual labelling is then utilized to retrieve fashionable goods or products. The manually performed annotation of the content of images is subjective and time-consuming. Different annotators add different descriptions to the same clothing item. In a similar way based on the situation at hand, a single individual may think about the same sight many times at various periods. 

To get the desired jewelry goods, content-based jewelry item retrieval (CBJIR) is often employed in online shopping platforms and searches on sites such as Taobao, Jingdong, Google, Baidu, and many more. Individuals adore photographing their daily surroundings and purchasing their favorite goods online \cite{corbiere2017leveraging,hadi2015buy}. Consumers may speedily find the selected jewellery products online with the help of the CBJIR methodology. However, research on complex fashion products such as jewellery items has little momentum due to the simultaneous nature of multiple jewellery products, different jewellery products' occlusion in images or visual streams, shape deformation, and the unavailability of appropriate datasets. According to industry studies, the scale of the local and worldwide fashion marketplaces is continually growing. Although the technology for obtaining images of fashion goods has progressed substantially over the last two decades, retrieving jewelry items in cross-domain situations still needs improvement. 

To overcome the above challenges, we proposed a content-based jewellery item retrieval system that uses the local region-based histograms in HSV color space to retrieve jewellery items better. Our novel Jewellery Classification Module (JCM) plays an integral part in extracting the specific features from the queried and jewellery databases. The JCM extracts the specific feature vectors from the query image using five local regions explained in section 3. The JCM is also applied to the jewellery databases to remove the feature vectors from the vast collection of items. Finally, the similarity score between the database and the query features is matched to retrieve the most relevant jewellery items from the database.

The rest of the article is arranged as follows; section \ref{relatedWork} reviews some recent related work of content-based fashion image retrieval systems. The complete description of the proposed Content-based jewellery item retrieval using the local region-based histograms is presented in section \ref{ProposedMethod}. The implementation of the proposed method and comparative analysis with the baseline methods are discussed in section \ref{experiments&results}. Finally, the conclusion of the article is presented in section \ref{conclusion}. 

\section{Related Work}
\label{relatedWork}
In recent decades, fashion research has advanced significantly, with the use of reference photos for fashion retrieval being among the most effective methodologies and a research hotspot. Every day, thousands of images and their associated electronic information are being added to online shopping web pages, worsening the fashion retrieval problem. Utilizing similarity measurement and feature extraction viewpoints, researchers have contributed to improving the precision of Fashion Item Retrieval (FIR) in current years \cite{gajic2018cross,park2019study}. Several artificial intelligence frameworks were developed in recent decades to improve FIR performance while maximizing computational resource utilization. FIR frameworks are categorized into two main parts, i.e. text-based fashion item retrieval systems and content-based fashion item retrieval systems. 

\begin{figure*}[!htb]
	\centering
	\includegraphics[width=\textwidth]{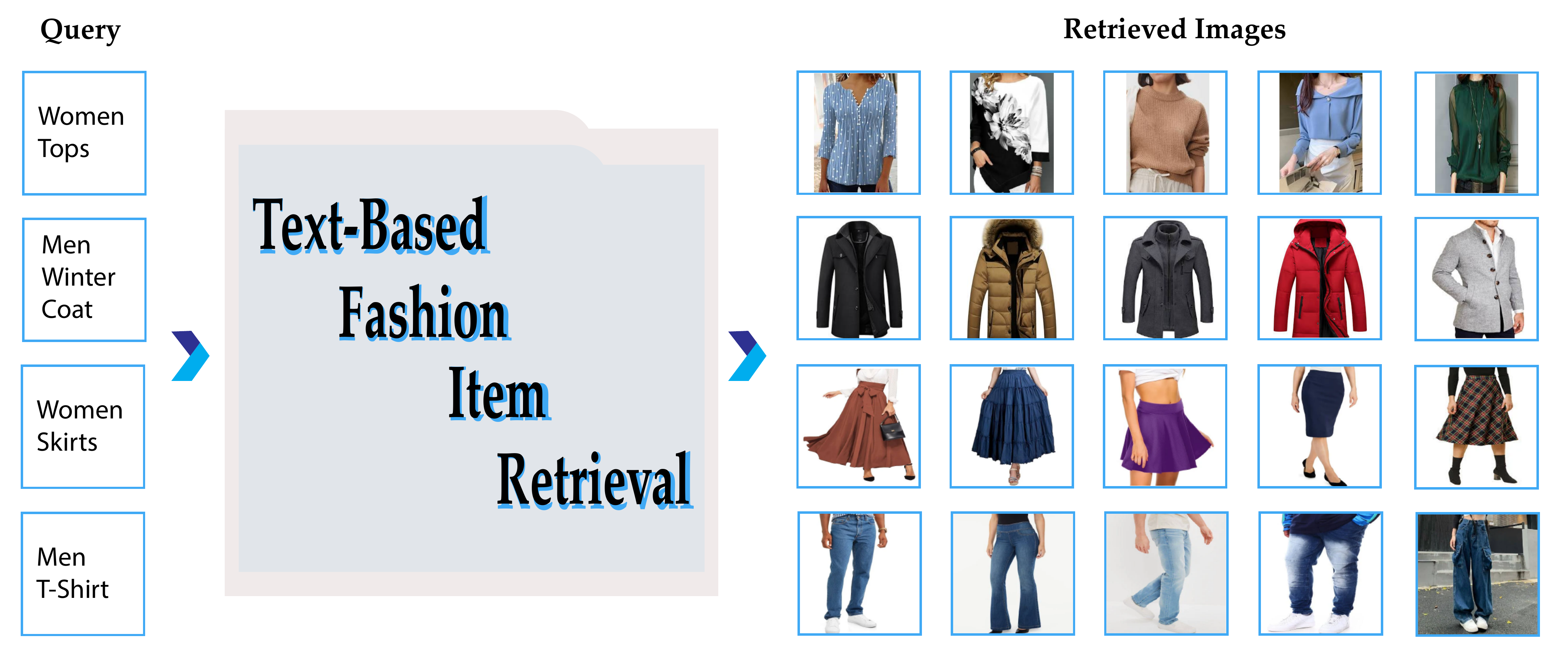}
	\caption{General outcome of Text-Based Fashion Item Retrieval methods using text/keywords query.}
	\label{img:TBFIR-RW}
\end{figure*}

\begin{figure*}[!htb]
	\centering
	\includegraphics[width=\textwidth]{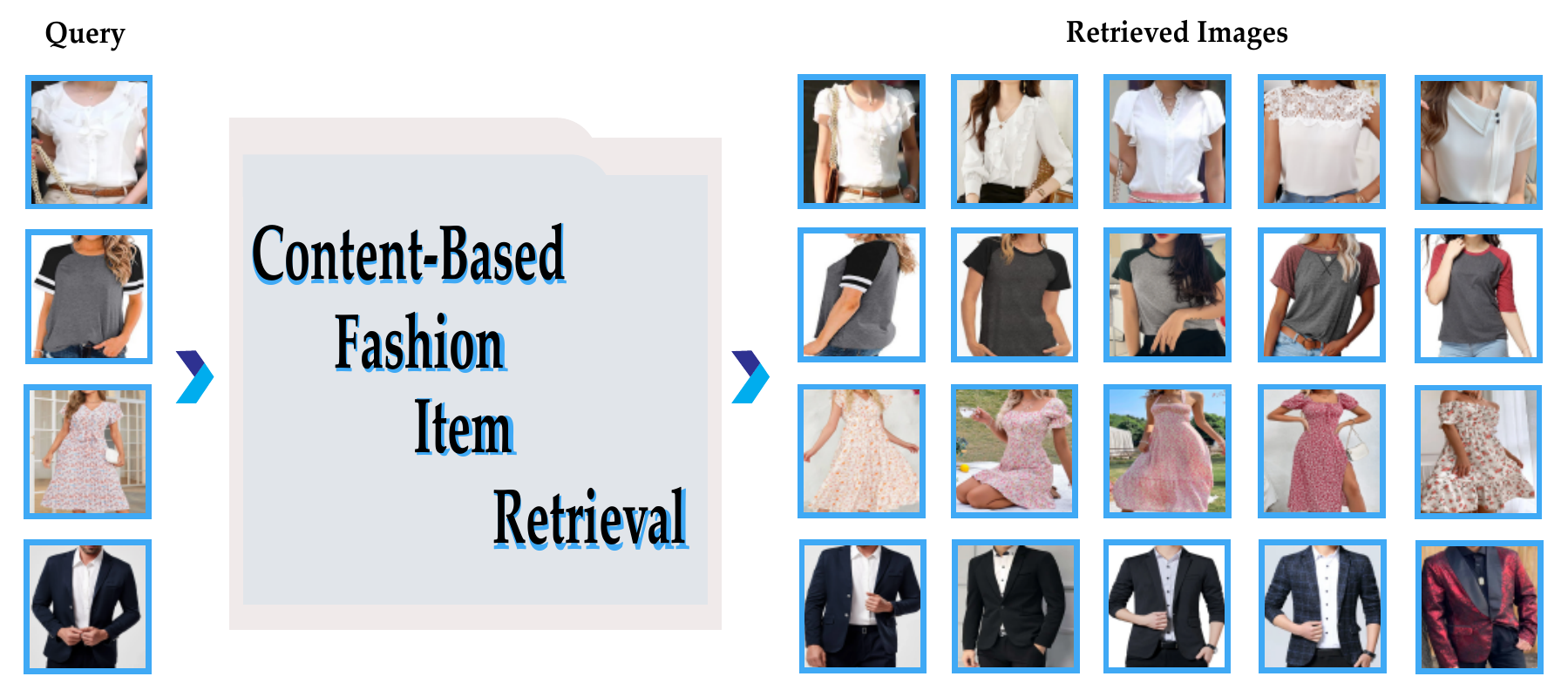}
	\caption{General outcome of Content-Based Fashion Item Retrieval methods using image query.}
	\label{img:CBFIR-RW}
\end{figure*}

\textbf{Text-based fashion items retrieval (TBFIR) systems:} The TBFIR systems retrieve the desired products for the consumers based on provided text or keyword queries. Text/Keyword-based fashion item retrieval systems have a simpler architecture than content-based retrieval systems. The approaches proposed for TBFIR systems range from a "simple frequency-of-occurrence-based scheme to an ontology-based scheme \cite{alkhawlani2015text}." TBFIR is more effective as compared to content-based FIR models in handling semantic inquiries. 

Previous unimodal-text retrieval systems relied heavily on manual picture annotations, where the user annotated the content of fashion items using keywords. Such manual comments are then utilized to retrieve the clothing items from the databases \cite{shoib2023methods}. The manually performed annotation of the content of the image is time-consuming and a subjective process. Different annotation experts add distinct explanations to the same clothing item. In a similar vein based on the situation at hand, a single individual may think about the same sight countless times during various periods \cite{li2011text}. As a result, annotations made manually can be used in particular domains such as virtual museums, online libraries, individual recordings, and many more. 

In text-based retrieval models, automatic image indexing could be an approach to this problem. Additionally, there are numerous autonomous indexing strategies, the most prominent of which is "count the frequency of occurrence of words." The actual distance between the words (used to describe the image) determines the weighting of the words. Moreover, reactions from customers to image results may be utilized to enhance keywords. The weighting technique for image keywords similarly heavily depends upon the domain. Many aesthetic features of fashion products are difficult to convey in terms. Figure \ref{img:TBFIR-RW} presents the general outcome of text-based fashion item retrieval methods using text or keywords as a query.

\begin{figure*}[h]
	\centering
	\includegraphics[width=\textwidth]{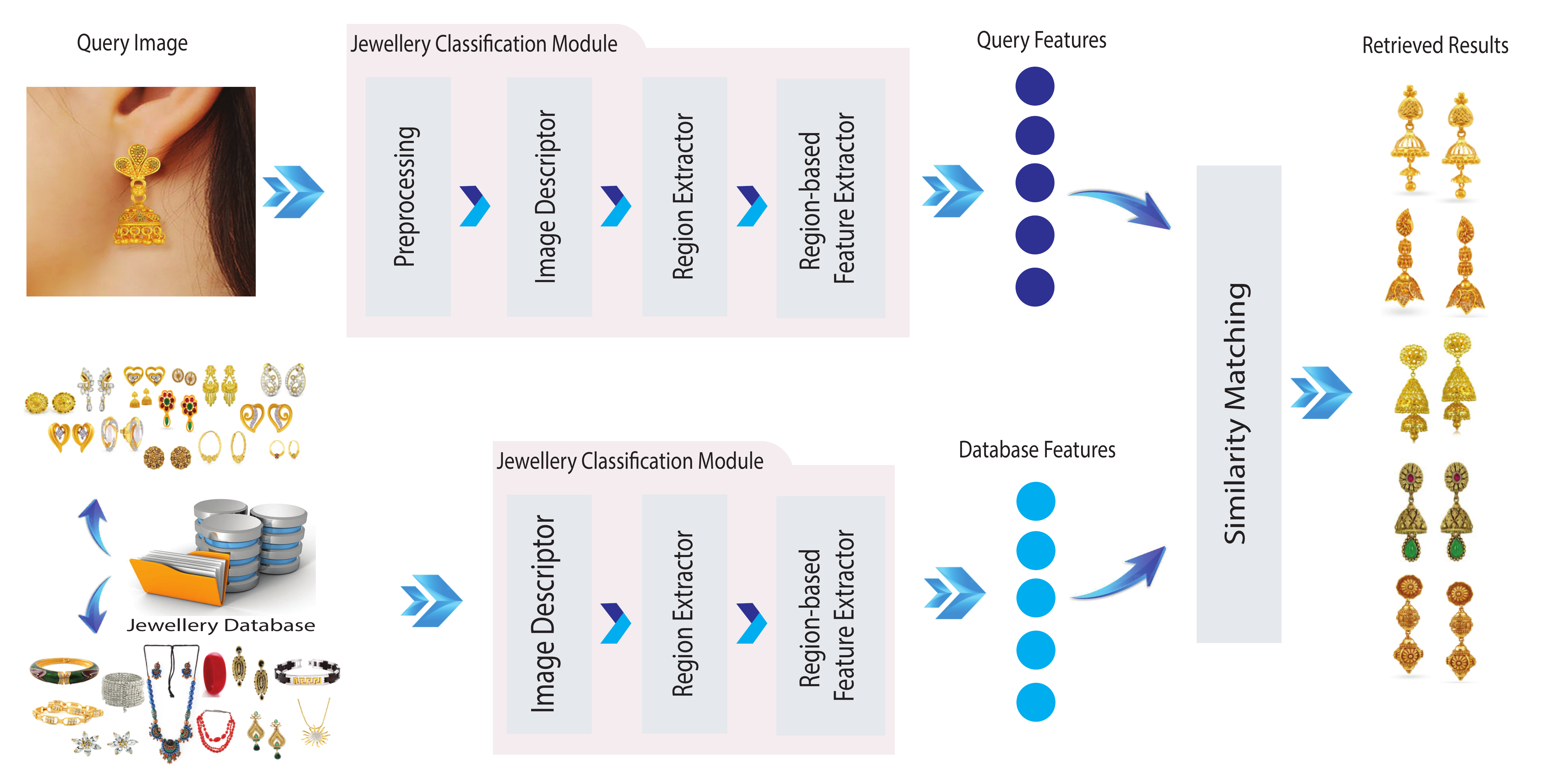}
	\caption{Flow diagram of proposed content-based jewellery item retrieval method.}
	\label{img:FD}
\end{figure*}

\textbf{Content-based fashion item retrieval (CBFIR) systems:} The CBFIR systems are used to retrieve the desired products for the consumer based on provided reference images\cite{adrakatti2016search}. In content-based retrieval systems, feature extraction strategies are critical for the retrieval of desired clothing items from the huge collection of images \cite{tai2023content,mustaffa2019dress,li2016retrieval}. Where the similarity and feature vectors (FV) are two essential mechanisms. In the CBFIR system, the features are extracted to represent the unique instance of queried and databased images. The extracted features are used for calculating database and reference image similarities. Utilizing similarity measurement and feature extraction approaches, researchers have contributed to increasing the precision of unimodal FIR in recent years. Several artificial intelligence frameworks have already been suggested in recent decades to improve FIR achievement while making the best use of computing power \cite{chang2022content}. As compared to TBFIR systems, using a reference image as an input in CBFIR enables customers to convert rich information regarding their preferred fashion product or item. 

Likewise, using a reference image as an input query to the retrieval system has unified features as compared to putting text/keywords as an input query. But if some consumers desire some additional attributes regarding the fashion item along with the queried reference image, then the content-based fashion item retrieval systems are not able to retrieve the desired product with these additional attributes. Figure \ref{img:CBFIR-RW} presents the general outcome of Content-based fashion item retrieval methods using reference images as a query.

\section{Proposed Method}
\label{ProposedMethod}

Content-based Jewellery item retrieval (CBJIR) method quantifies the content of images. Mostly, consumers use jewelry item retrieval systems to search for their desired jewellery products or items on online marketplaces using a sample query reference image. Considering recent developments, due to the simultaneous nature of various jewelry items, jewelry goods' occlusion in images or visual streams, as well as shape deformation, CBJIR still has limitations whenever it pertains to visual searching in the actual world. To overcome these limitations, in this research we proposed a region-based jewellery classification module using local histograms to extract the specific feature vectors from the queried image to retrieve desired jewellery items. The general structure diagram of the proposed content-based jewellery item retrieval method is presented in Figure \ref{img:FD}. We provide the cropped jewellery item image to the jewellery classification module to retrieve selected jewellery products. In the preprocessing stage, we first convert the RGB space of the query image into the 3D HSV color space. RGB values are used to quantify the simple values of an image and cannot impressionist the human-style perception of colors. We designed a specific 3D HSV color space to illustrate human-style color perception better. In the jewellery classification module, the image descriptor optimally selects the number of bins for 3D HSV color space histograms. The quantization of pixel intensities of a query image is elaborated with these histogram bins. We optimally select the number of bins for 3D HSV color space histograms. To yield the maximum dimensions of feature vectors, we choose 10 bins for the hue channel, 3 bins for the value channel, and 14 bins for the saturation channel. 

\begin{figure}[h]
	\centering
	\includegraphics[width=0.5\textwidth, height=80mm]{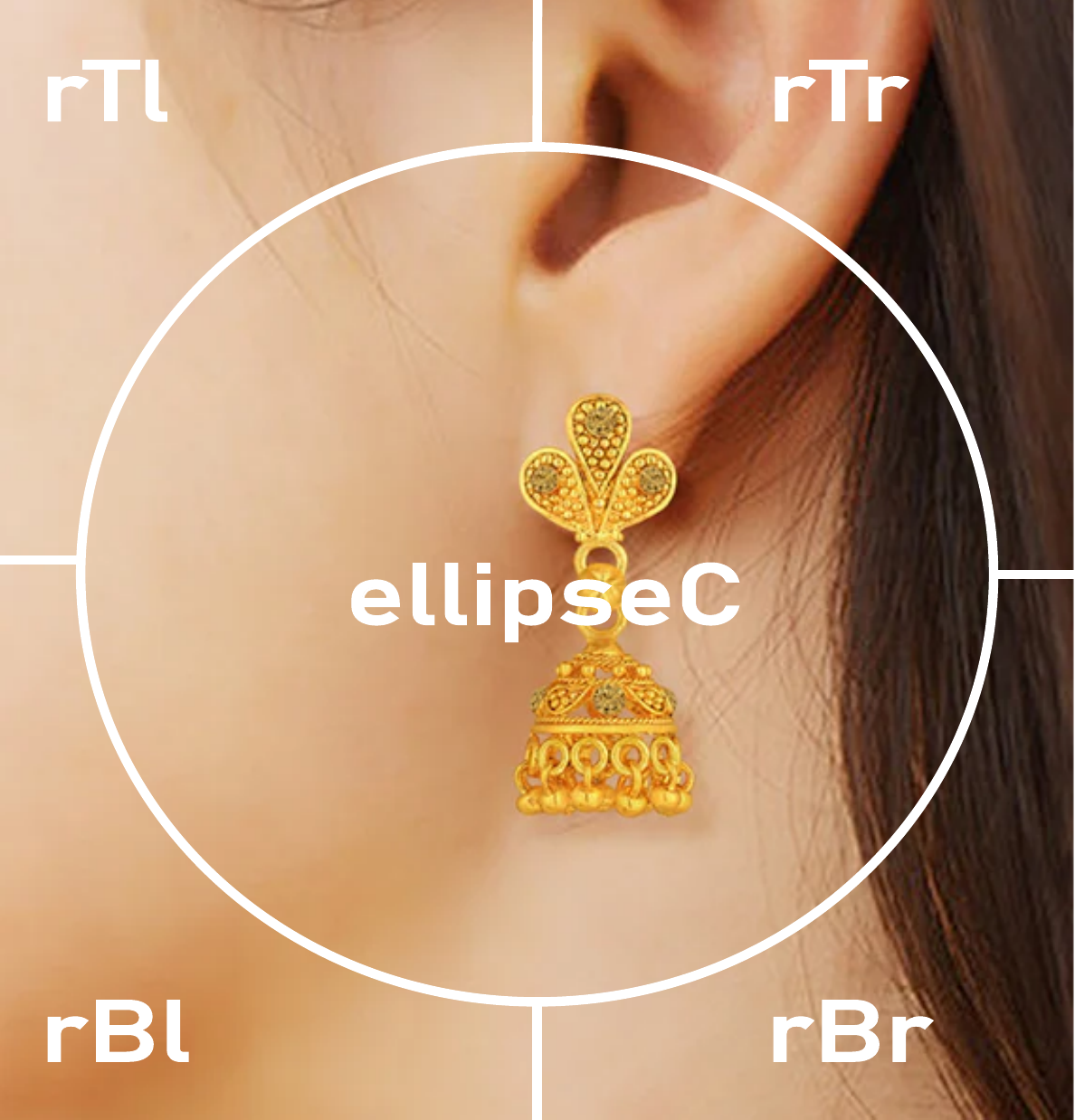}
	\caption{Local region-based division for 3D HSV color space, where $rTl$ = top-left region, $rTr$ = top-right region, $rBl$ = bottom-left region, $rBr$ = bottom-right region, and $ellipseC$ = central region.}
	\label{img:local-regions}
\end{figure}

We proposed a novel region extractor in the jewellery classification module to compute the local region-based histogram for 3D HSV color space. We add the local region-based histogram rather than the global histogram of the entire queried image to calculate the locality while extracting feature vectors. To calculate the locality of the queried image and database images, we partitioned the images into the top left and right, bottom left and right, and central regions. The Jewellery classification module used region-based localities to extract the optimum feature vectors from the query image to enhance the retrieval outcome of jewellery items. Figure \ref{img:local-regions} represents the division of $rTl, rTr, rBl, rBr,$ and $ellipseC$ region of queried image. We can calculate these regions by using four indexes of the input image, i.e. startX, endX, startY, and endY. The $rTl, rTr, rBl, rBr,$ and $ellipseC$ are calculated using Equations \ref{eq:topl,topr,br,bl} and \ref{eq:ellipse}, simultaneously. 

\begin{equation}
	\begin{gathered}
		rTl = (\theta,cX,\theta,cY)\\
		rTr = (cX,w,\theta,cY)\\
		rBl = (\theta,cX,cY,h)\\
		rBr = (cX,w,cY,h)\\
		region = [(rTl),(rTr),(rBr),(rBl)]\\
	\end{gathered}	
\label{eq:topl,topr,br,bl}
\end{equation}

Where $rTl, rTr, rBr,$ and $rBl$ contain the input image's starting and ending index of $(X,Y)$-coordinate. The value of $\theta$ is equal to zero here to indicate the starting point of $(X,Y)$-coordinate, $w$ is the width and $h$ is the height of an image, and $(cX,cY)$ represents the center $(X,Y)$-coordinate.

In this jewellery classification module, we calculate the center region of the queried image with the perspective of an ellipse. The construction of the ellipse is represented in Equation \ref{eq:ellipse}. 

\begin{equation}
	\begin{gathered}
		r = (int(w*0.7)/2, int(h*0.7)/2)\\
		ellipseC = (r,(cX,cY), (axesX,axesY),0,0,360,255,-1) 
	\end{gathered}	
	\label{eq:ellipse}
\end{equation}

Where $(axesX,axesY)$ represents the length of the ellipse and $r$ is the radius of ellipse. To optically extract feature vectors of jewellery items, we use a radius of the ellipse equal to 70 percent of the height and width of the queried image. 0 and 360 are starting and ending angles of an ellipse. 255 represents the color of an ellipse, and -1 represents the border size.

We loop over all five local regions of the queried image and construct a mask for each region separately to extract features from it. We update the feature vector list by lopping over the rTl, rTr, rBl, rBr, and ellipse region of the image. The local histograms of the queried image are calculated using the explained number of bins for HSV space. Similarly, feature vectors of all the database images are calculated using the exact mechanism through the jewellery classification module. 
Chi-square distance calculates the similarities and dissimilarities between region-based histogram bins. The similarity matching between query and database features is calculated using chi-square distance using Equation \ref{eq:distance}.

\begin{equation}
	\label{eq:distance}
	X^{2}(xbin,ybin) = \frac{1}{2} \sum_{a=1}^k \frac{(xbin_{a}-ybin_{a})^{2}}{xbin_{a}+ybin_{a}}
\end{equation}

Where xbin and ybin calculate similarities between the database and query feature bins. After similarity matching, the closest jewellery retrieval results from the database are retrieved. The proposed method is tested on the two well-known jewellery image retrieval datasets, which details are provided in the section \ref{experiments&results}.

\section{Experiments and Results}
\label{experiments&results}

We performed various sorts of experiments to assess the efficiency of the proposed methodology. On the RingFIR \cite{islam2021ringfir} and Fashion Product Images \cite{aggarwal2019fashion} datasets, we have contrasted the retrieval accuracy of the suggested approach to that of other mentioned retrieval strategies. The following demonstration shows the complete detail of experimental configuration, setup, and results on the RingFIR and Fashion Product Images dataset. The performance of the content-based jewellery item retrieval using the local region-based histogram methods experiments are evaluated using the Top-k accuracy. Top-k accuracy measures the retrieval accuracy of relevant retrieved images from the database. Our experiments compute the top 1, 5, 10, 15, and 20 retrieval accuracy on the RingFIR and Fashion Product Images dataset. 

\subsection{Datasets}
\begin{figure}[h]
	\centering
	\includegraphics[width=0.5\textwidth, height=50mm]{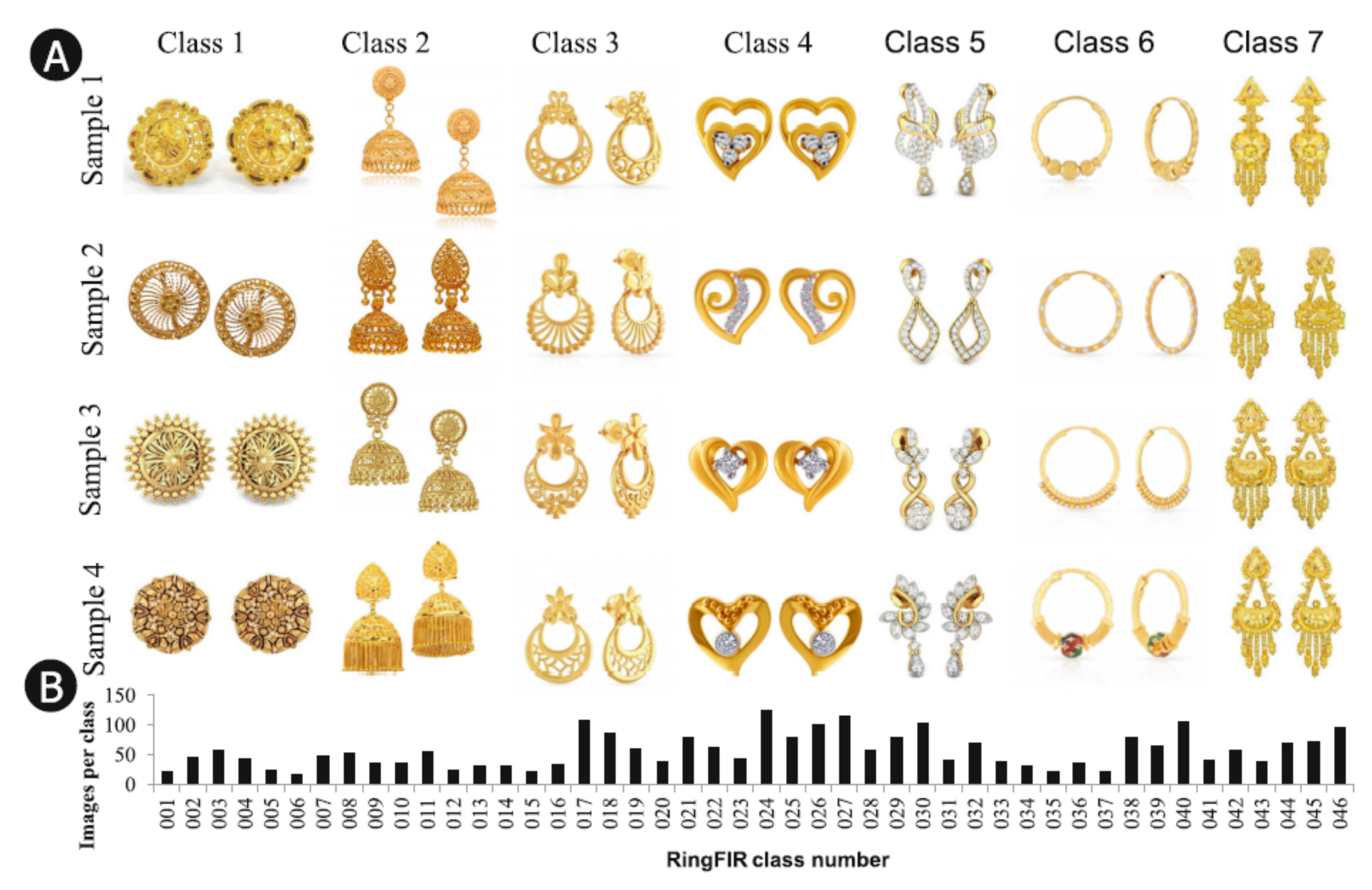}
	\caption{RingFIR dataset: A) random sample examples of different classes, B) Images distribution in different classes \cite{islam2021ringfir}}
	\label{img:DS-RingFIR}
\end{figure}

\textbf{RingFIR:}
The RingFIR \cite{islam2021ringfir} is a diversified collection of earrings from various jewellery catalogues. This dataset consists of approximately 2,651 high-quality different images of golden earrings. This collection of different images is structurally categorized into 46 classes according to their patterns, design, and structures. The RingFIR is one of the most suitable datasets for our proposed method for retrieval of desired jewellery items from the database. Figure \ref{img:DS-RingFIR} presents the random sample images of the RingFIR dataset. 

\begin{figure}[h]
	\centering
	\includegraphics[width=0.5\textwidth, height=50mm]{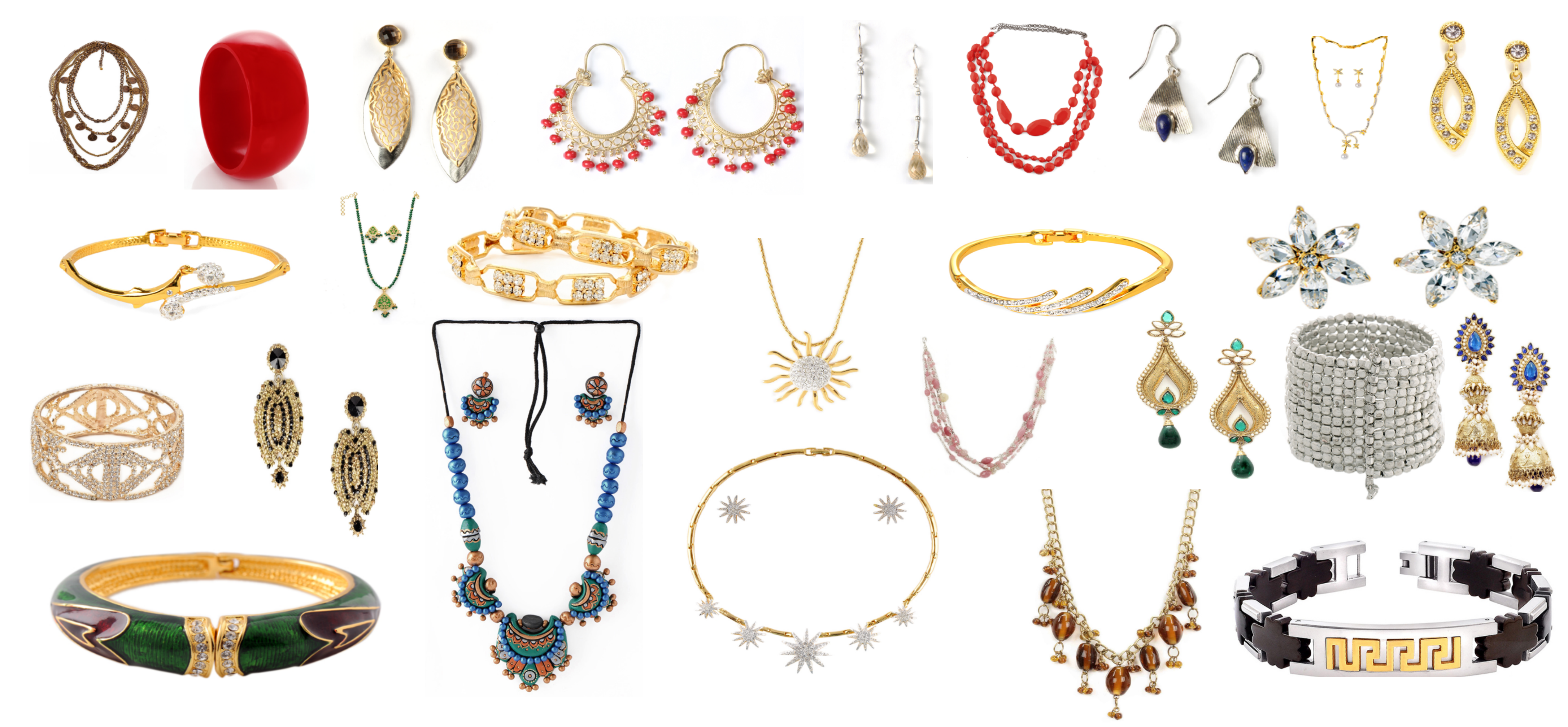}
	\caption{Random sample examples of different Fashion Product Images dataset \cite{aggarwal2019fashion}.}
	\label{img:Ds-FashionImagesProduct}
\end{figure}

\begin{figure*}[!htb]
	\centering
	\includegraphics[width=150mm, height=105mm]{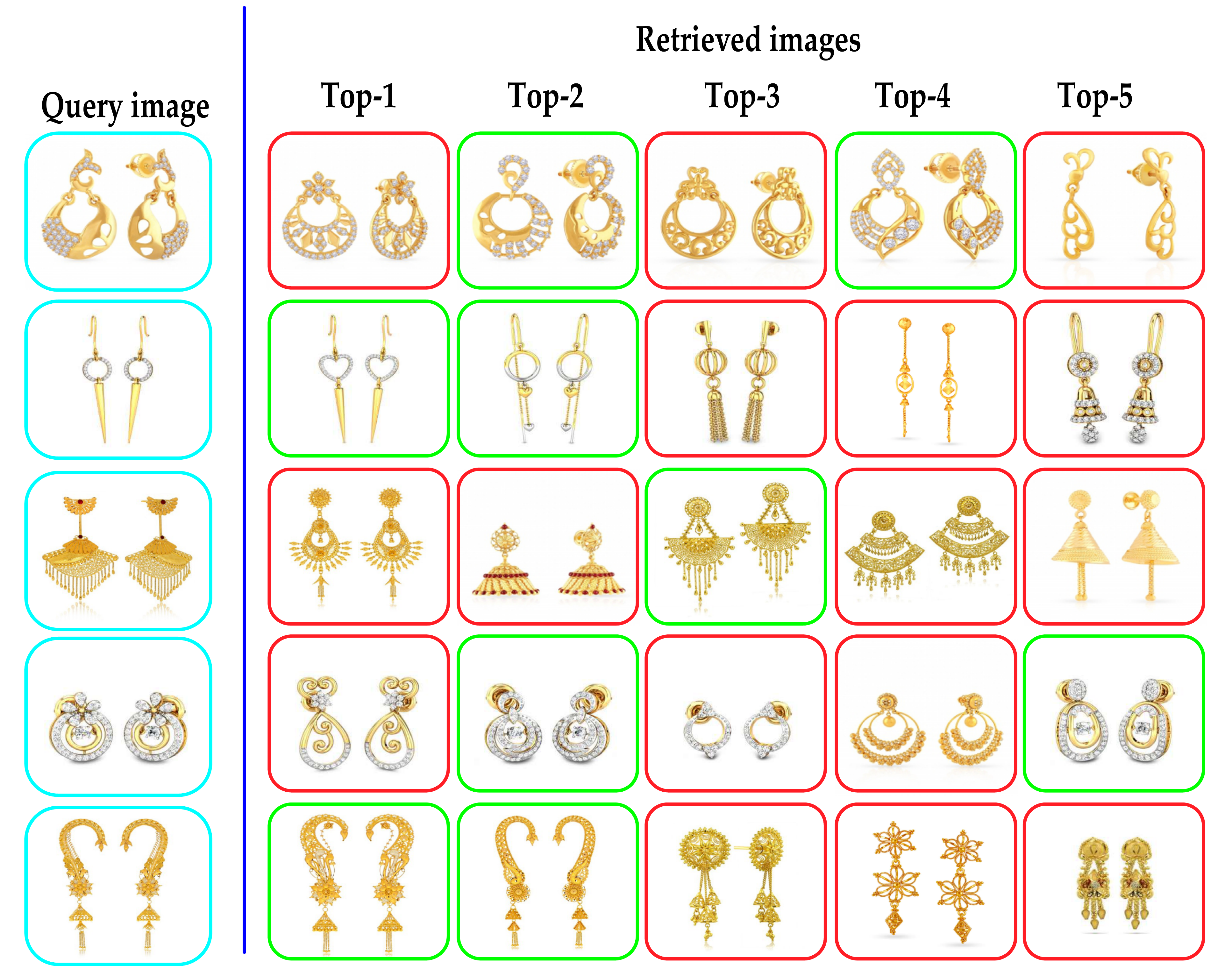}
	\caption{RingFIR retrieved results from the test split, where the green box presents successful retrieval and the red box presents unsuccessful retrieval.}
	\label{img:results-ringFIR-1}
\end{figure*}

\begin{figure*}[!htb]
	\centering
	\includegraphics[width=150mm, height=105mm]{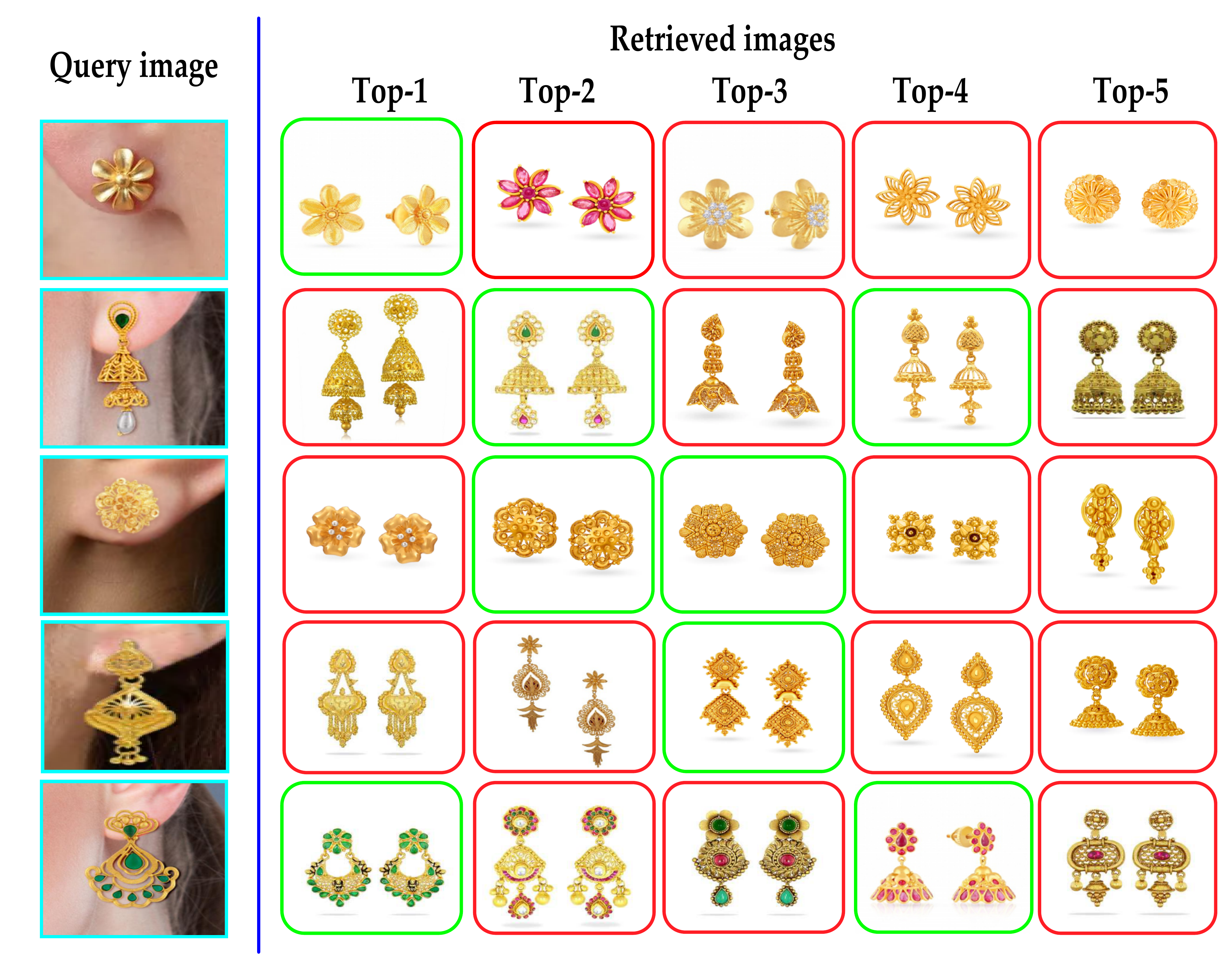}
	\caption{RingFIR retrieved results from the test split, where the green box presents successful retrieval and the red box presents unsuccessful retrieval.}
	\label{img:results-ringFIR-2}
\end{figure*}

\textbf{Fashion Product Images dataset:} 
The Fashion Product Images dataset \cite{aggarwal2019fashion} contains 44,441 number of images along with a 3-class hierarchy. Additionally, the dataset contains a masterCategory and subCategory levels with 4 and 21 classes simultaneously. This dataset contains mixed jewellery items like necklaces, earrings, bracelets, rings, etc. To assess the performance of the proposed Content-based jewellery item retrieval using the local region-based histograms method we extract all the jewellery item images from the vast collection of Fashion Product Images dataset. A total of 1,081 different jewellery item images are available in this dataset which we used for the retrieval task of the proposed method. Figure \ref{img:Ds-FashionImagesProduct} presents the random sample images of the RingFIR dataset.

\subsection{Experimental setup and results:}

The proposed method experiments on the referred above datasets are conducted on Intel(R) Core(TM) i7-9750H CPU, 32GB-RAM system with NVIDIA GeForce GTX 1660Ti GPU. Anaconda is used for the development environment with Pillow, Flask, TensorFlow, Keras, and other libraries. In the proposed method, the jewellery classification module is used to extract the specific feature vectors from the queried image, which are then used for similarity matching of extracted database features. The region and region-based feature extractors are vital in extracting specific jewellery item features from the query image and jewellery databases. 

The following demonstrations show the detailed results and effectiveness of this work.
The performance of the proposed content-based jewellery item retrieval using the local region-based histograms method is first evaluated on the RingFIR dataset, where we used a 90:10 percent ratio for training and testing of images for the retrieval of jewellery items. For better evaluation of results, we perform experiments on test split of the RingFIR dataset and some experiments on the cropped jewellery items images from out-of-database sources. Figure \ref{img:results-ringFIR-1} shows the retrieved images results of the test split on the RingFIR dataset. Query images are taken from the test split of the RingFIR dataset which is presented on the left side of the figure and the right side shows the top 1, 2, 3, 4, and 5 retrieved images results. Similarly, Figure \ref{img:results-ringFIR-2} shows the retrieved images of out-of-database queried photos, where cropped query images of jewellery items are taken from the miscellaneous web sources presented on the left side of the figure. The right side presents the top 1, 2, 3, 4, and 5 retrieved image results. 

The performance of the content-based jewellery item retrieval using the local region-based histograms method’s experiments is evaluated using the Top-k accuracy. Where, we compute the top 1, 5, 10, 15, and 20 retrieval accuracy on the RingFIR dataset. The quantitative retrieval results of the baseline methods and the proposed method on the test split of the RingFIR dataset are presented in Table \ref{tab:results-RingFIR-1}. The experimental results clearly show the improvement of the proposed method over the baseline methods. Our proposed method obtains 32.67, 59.31, 74.24, 78.18, and 90.18 top 1, 5, 10, 15, and 20 respectively. The results clearly show improvement in top 1, 5, 10, and 20 retrieval accuracy over the baseline methods except for the top-15 accuracy of the DSSN \cite{islam2023dssn} method. Additionally, the quantitative retrieval results of the baseline methods and the proposed method from the out-of-database reference image on the RingFIR dataset are presented in Table \ref{tab:results-RingFIR-2}. The experimental results clearly show the improvement of the proposed method over the baseline methods. Our proposed method obtains 22.67, 56.26, 72.19, 75.51, and 85.25 top 1, 5, 10, 15, and 20 respectively. The results clearly show improvement in the top 5, 10, and 20 retrieval accuracy over the baseline methods except for the top 1 and 15 accuracy of the DSSN \cite{islam2023dssn} method.

\begin{table}[]
	\centering
	\scriptsize 
	\caption{Baseline methods and the proposed method’s top 1, 5, 10, 15, and 20 retrieved results on the test split of the RingFIR dataset, where bold presents the best results.}
	\label{tab:results-RingFIR-1}
	\begin{tabular}{p{2.8cm}p{0.6cm}p{0.6cm}p{0.6cm}p{0.6cm}p{0.6cm}}
		\bottomrule
		\rowcolor[HTML]{98DAED}
		\textbf{Method}   & \textbf{Top 1} & \textbf{Top 5} & \textbf{Top 10} & \textbf{Top 15} & \textbf{Top 20} \\
		\toprule
		RingFIR-VGG16 \cite{ha2018image} & 15.22 & 39.13 & 56.52 & 65.22 & 71.74 \\
		RingFIR-ResNet50 \cite{pelka2018annotation} & 15.21 & 28.26 & 43.47 & 54.68 & 61.23 \\
		RingFIR-MobileNet \cite{ilhan2020fully} & 10.87 & 32.61 & 43.48 & 53.58 & 58.28 \\
		EfficentNetV2 \cite{tan2021efficientnetv2} & 6.52  & 26.08 & 36.95 & 50.0  & 60.86 \\
		DSSN \cite{islam2023dssn} & 26.74 & 57.17 & 72.17 & \textbf{79.57} & 85.87 \\
		Ours              & \textbf{32.67} & \textbf{59.31} & \textbf{74.24}  & 78.18 & \textbf{90.48} \\
		\bottomrule
	\end{tabular}
\end{table}

\begin{table}[]
	\centering
	\scriptsize
	\caption{Baseline methods and the proposed method’s top 1, 5, 10, 15, and 20 retrieved results from the out-of-database reference image on the RingFIR dataset, where bold presents the best results.}
	\label{tab:results-RingFIR-2}
	\begin{tabular}{p{2.8cm}p{0.6cm}p{0.6cm}p{0.6cm}p{0.6cm}p{0.6cm}}
		\bottomrule
		\rowcolor[HTML]{98DAED}
		\textbf{Method}   & \textbf{Top 1} & \textbf{Top 5} & \textbf{Top 10} & \textbf{Top 15} & \textbf{Top 20} \\
		\toprule
		RingFIR-VGG16 \cite{ha2018image} & 13.64 & 33.48 & 48.64 & 62.15 & 70.32 \\
		RingFIR-ResNet50 \cite{pelka2018annotation} & 12.97 & 29.23 & 40.73 & 53.32 & 58.35 \\
		RingFIR-MobileNet \cite{ilhan2020fully} & 8.21 & 30.59 & 41.62 & 51.22 & 60.06 \\
		EfficentNetV2  \cite{tan2021efficientnetv2}   & 5.58 & 24.37 & 32.59 & 47.15 & 57.24 \\
		DSSN  \cite{islam2023dssn}  & \textbf{23.03} & 54.18 & 68.41 & \textbf{80.57} & 82.68  \\
		Ours              & 22.67 & \textbf{56.26} & \textbf{72.19} & 75.71 & \textbf{85.25} \\
		\bottomrule
	\end{tabular}
\end{table}

\begin{figure*}[!htb]
	\centering
	\includegraphics[width=150mm, height=103mm]{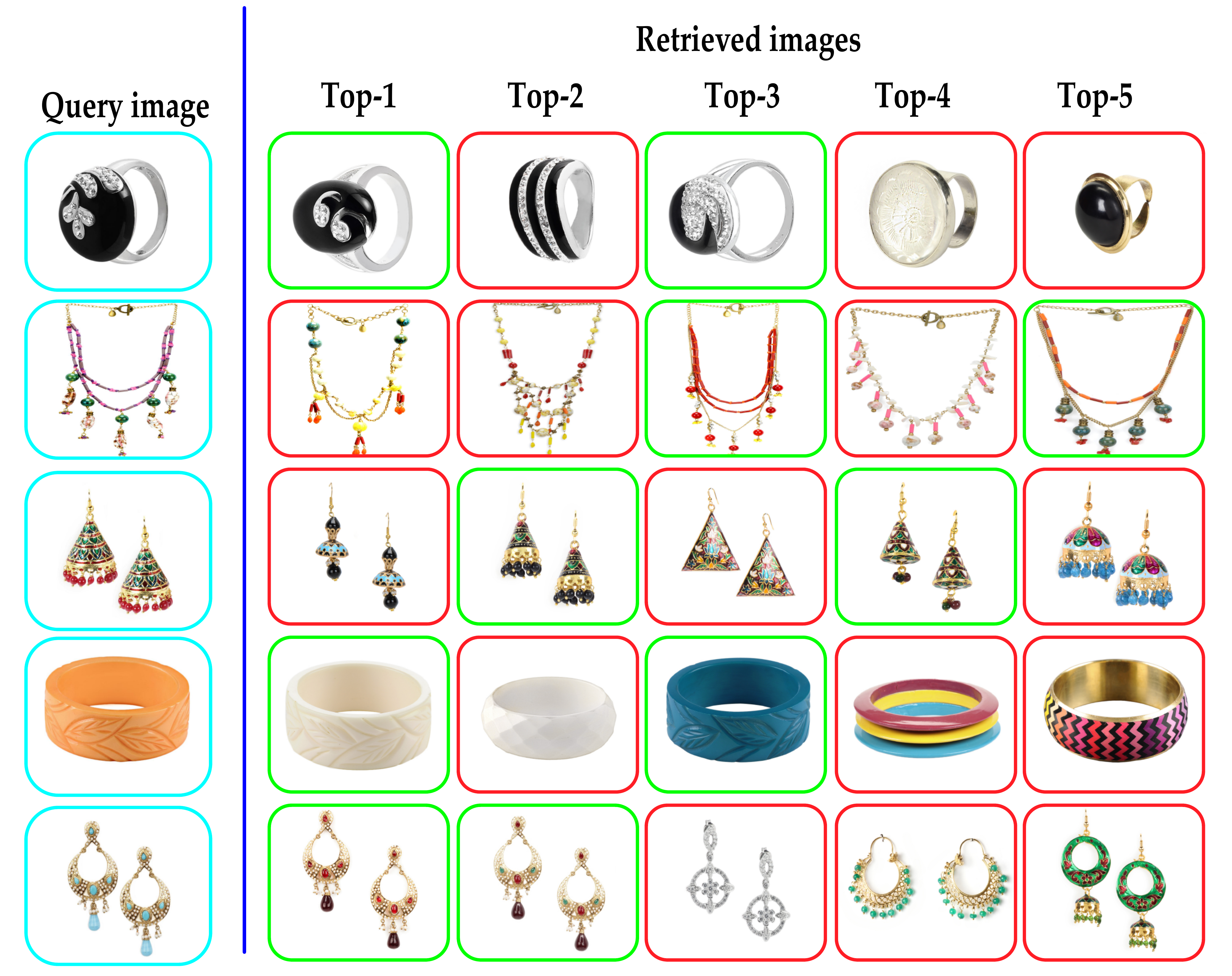}
	\caption{Fashion Product Images dataset retrieved results from the test split, where the green box presents successful retrieval and the red box presents unsuccessful retrieval.}
	\label{img:results-FashionProductImages-1}
\end{figure*}

\begin{figure*}[!htb]
	\centering
	\includegraphics[width=150mm, height=103mm]{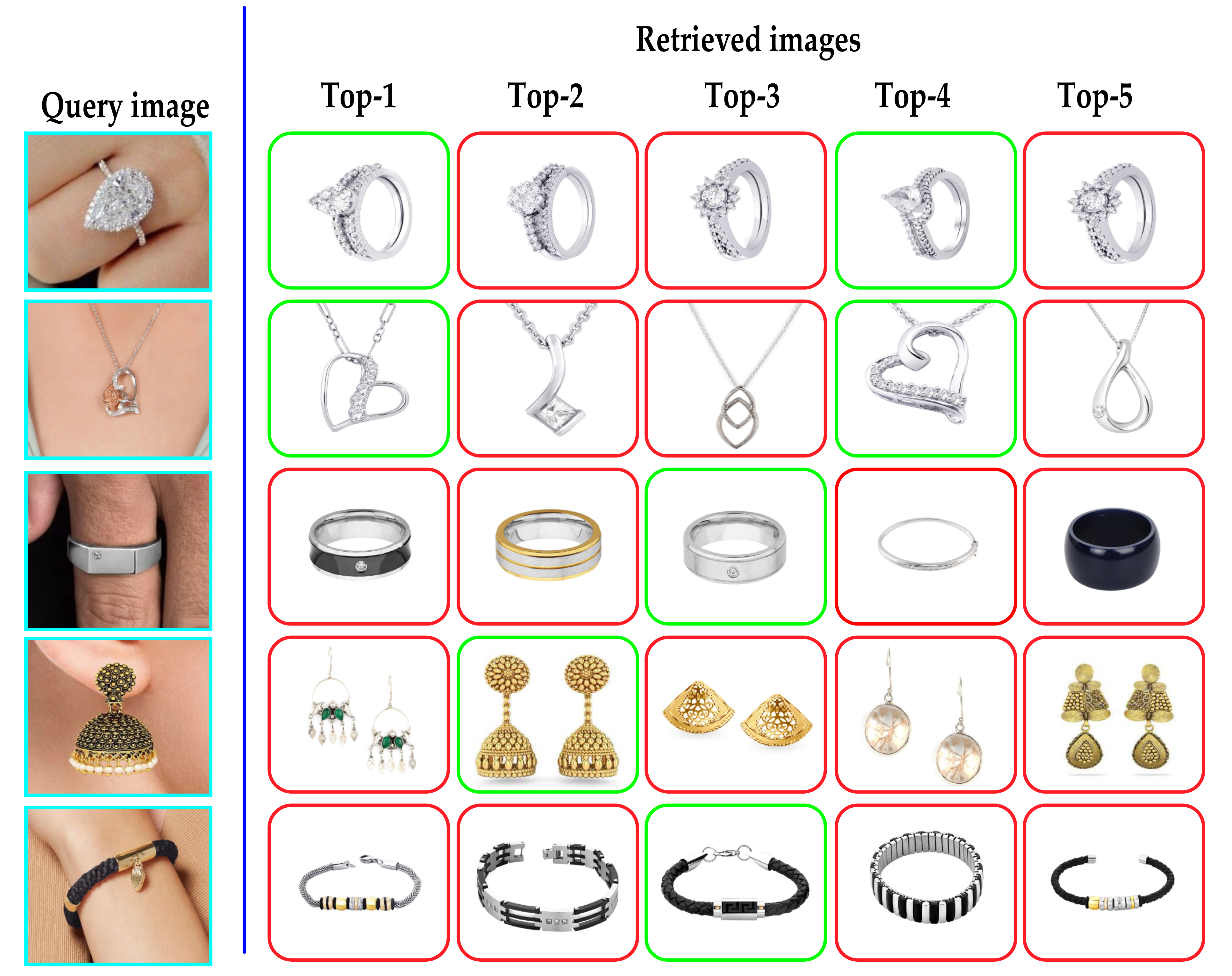}
	\caption{Fashion Product Images dataset retrieved results from the test split, where the green box presents successful retrieval, and the red box presents unsuccessful retrieval.}
	\label{img:results-FashionProductImages-2}
\end{figure*}

Additionally, the proposed method is also evaluated on the Fashion Product Images dataset, where we similarly used a 90:10 percent ratio for training and testing images for the retrieval of jewellery items. Firstly, we perform experiments on the test split of the Fashion Product Images dataset and then perform additional experiments on the cropped jewellery items images from out-of-database sources. Figure \ref{img:results-FashionProductImages-1} shows the retrieved images results of the test split on the Fashion Product Images dataset. Query images are taken from the test split of the Fashion Product Images dataset which is presented on the left side of the figure and the right side shows the top 1, 2, 3, 4, and 5 retrieved images results. Similarly, Figure \ref{img:results-FashionProductImages-2} shows the retrieved images results of the out-of-database queried images on the Fashion Product Images dataset, where cropped query images of jewellery items are taken from the miscellaneous web sources presented on the left side of the figure and the right side presents the top 1, 2, 3, 4, and 5 retrieved images results.

The quantitative retrieval results of the baseline methods and the proposed method on the test split of the Fashion Product Images dataset are presented in Table \ref{tab:results-FashionProductImages-1}. The experimental results clearly show the improvement of the proposed method over the baseline methods. Our proposed method obtains 27.29,	60.36, 69.08, 80.19, and 88.28 top 1, 5, 10, 15, and 20 respectively. The results clearly show improvement in top 1, 5, 15, and 20 retrieval accuracy over the baseline methods except for the top 10 accuracy of the DSSN \cite{islam2023dssn} method. Additionally, the quantitative retrieval results of the baseline methods and the proposed method from the out-of-database reference image on the Fashion Product Images dataset are presented in Table \ref{tab:results-FashionProductImages-2}. The experimental results clearly show the improvement of the proposed method over the baseline methods. Our proposed method obtains 20.58, 58.31, 70.57, 81.52, and 91.03 top 1, 5, 10, 15, and 20 respectively. The results clearly show improvement in top 1, 5, 10, 15, and 20 retrieval accuracy over the baseline methods.

\begin{table}[]
	\centering
	\scriptsize
	\caption{Baseline methods and the proposed method’s top 1, 5, 10, 15, and 20 retrieved results on the test split of the Fashion Product Images dataset, where bold presents the best results.}
	\label{tab:results-FashionProductImages-1}
	\begin{tabular}{p{2.8cm}p{0.6cm}p{0.6cm}p{0.6cm}p{0.6cm}p{0.6cm}}
		\bottomrule
		\rowcolor[HTML]{98DAED}
		\textbf{Method}   & \textbf{Top 1} & \textbf{Top 5} & \textbf{Top 10} & \textbf{Top 15} & \textbf{Top 20} \\
		\toprule
		RingFIR-VGG16 \cite{ha2018image} & 12.34 & 32.85 & 49.55 & 63.51 & 69.58 \\
		RingFIR-ResNet50 \cite{pelka2018annotation} & 12.52 &	25.12 &	37.86 &	52.85 &	60.41  \\
		RingFIR-MobileNet \cite{ilhan2020fully} & 7.68  &	29.81 &	39.62 &	50.21 &	55.25 \\
		EfficentNetV2 \cite{tan2021efficientnetv2} & 4.12  &	19.58 &	32.25 &	47.32 &	58.75 \\
		DSSN   \cite{islam2023dssn}  & 21.95 &	53.45 &	\textbf{69.58} & 76.85 & 82.62\\
		Ours              & \textbf{27.29} &	\textbf{60.36} &	69.08 &	\textbf{80.19} &	\textbf{88.28} \\
		\bottomrule
	\end{tabular}
\end{table}

\begin{table}[]
	\centering
	\scriptsize
	\caption{Baseline methods and the proposed method’s top 1, 5, 10, 15, and 20 retrieved results from the out-of-database reference image on the Fashion Product Images dataset, where bold presents the best results.}
	\label{tab:results-FashionProductImages-2}
	\begin{tabular}{p{2.8cm}p{0.6cm}p{0.6cm}p{0.6cm}p{0.6cm}p{0.6cm}}
		\bottomrule
		\rowcolor[HTML]{98DAED}
		\textbf{Method}   & \textbf{Top 1} & \textbf{Top 5} & \textbf{Top 10} & \textbf{Top 15} & \textbf{Top 20} \\
		\toprule
		RingFIR-VGG16 \cite{ha2018image} & 10.81 &	29.12 &	51.08 &	59.26 &	72.56  \\
		RingFIR-ResNet50 \cite{pelka2018annotation} & 11.75 &	31.64 &	42.26 &	60.11 &	64.25 \\
		RingFIR-MobileNet \cite{ilhan2020fully} & 9.68 &	27.42 &	39.72 &	52.85 &	59.93 \\
		EfficentNetV2 \cite{tan2021efficientnetv2} & 4.69 &	20.05 &	34.51 &	50.25 &	63.52 \\
		DSSN   \cite{islam2023dssn}  & 19.82 &	49.88 &	64.01 &	73.62 &	84.68   \\
		Ours              & \textbf{20.58} &	\textbf{58.31} &	\textbf{70.57} &	\textbf{81.52} &	\textbf{91.03} \\
		\bottomrule
	\end{tabular}
\end{table}

\section{Conclusion}
\label{conclusion}

A content-based jewellery item retrieval method using the local region-based histograms in HSV color space is proposed to achieve better accuracy. The core part of the proposed method is the jewellery classification module, which extracts the localities of jewellery items from the queried image based on five regions. The retrieval accuracy of the proposed method also improves by applying this jewellery classification module to the jewellery databases to extract the feature vectors optimally. The experimental results on the RingFIR and the Fashion Product Images dataset show the dominance of the proposed method over the baseline methods. This CBJIR method has significantly impacted the fashion industry to retrieve the desired jewellery items for their outfits. Still, there needs to be more databases available publicly to make the retrieval predictions better. In the future, a vast collection of jewellery item databases will be required to improve the retrieval accuracy of the CBJIR methods.

\section*{Declarations}

\subsection*{Conflict of interest}
The authors declare that they have no conflict of interest. The authors also declare that they have no known competing financial interests or personal relationships that could have appeared to influence the work reported in this paper



%
%

\bibliographystyle{IEEEbib}      
\bibliography{ref}   

%
%

\end{document}